\documentclass[runningheads]{llncs}
\usepackage{graphicx}

\usepackage{enumitem}
\usepackage{booktabs}
\usepackage{todonotes}

\usepackage{orcidlink}
\usepackage[sort,nocompress]{cite}

\usepackage{subcaption}
\usepackage{graphicx}

\usepackage{caption} 
\begin{document}
\title{DocILE 2023 Teaser: Document Information Localization and Extraction}

\author{Štěpán Šimsa\inst{1}\orcidlink{0000-0001-6687-1210} \and
Milan Šulc  \inst{1}\orcidlink{0000-0002-6321-0131} \and
Matyáš Skalický \inst{1}\orcidlink{0000-0002-0197-7134} \and\\
Yash Patel \inst{2}\orcidlink{0000-0001-9373-529X} \and
Ahmed Hamdi \inst{3}\orcidlink{0000-0002-8964-2135}
}
\authorrunning{Š. Šimsa et al.}
\institute{Rossum.ai, Czech Republic, \url{http://www.rossum.ai/}
\and Visual Recognition Group, Czech Technical University in Prague, Czech Republic
\and University of La Rochelle, France \\ 
\email{\{stepan.simsa,milan.sulc,matyas.skalicky\}@rossum.ai,\\ patelyas@fel.cvut.cz, ahmed.hamdi@univ-lr.fr}
}

\maketitle
\begin{abstract}
The lack of data for information extraction (IE) from semi-structured business documents is a real problem for the IE community. Publications relying on large-scale datasets use only proprietary, unpublished data due to the sensitive nature of such documents. Publicly available datasets are mostly small and domain-specific. The absence of a large-scale public dataset or benchmark hinders the reproducibility and cross-evaluation of published methods. The DocILE 2023 competition, hosted as a lab at the CLEF 2023 conference and as an ICDAR 2023 competition, will run the first major benchmark for the tasks of \emph{Key Information Localization and Extraction} (KILE) and \emph{Line Item Recognition} (LIR) from business documents. With thousands of annotated real documents from open sources, a hundred thousand of generated synthetic documents, and nearly a million unlabeled documents, the DocILE lab comes with the largest publicly available dataset for KILE and LIR. We are looking forward to contributions from the Computer Vision, Natural Language Processing, Information Retrieval, and other communities. The data, baselines, code and up-to-date information about the lab and competition are available at \url{https://docile.rossum.ai/}.

\end{abstract}

\section{Introduction}
The majority of business-to-business (B2B) communication takes place through the exchange of semi-structured documents such as invoices, purchase orders, and delivery notes. Information from the documents is typically extracted by humans and entered into information systems. This process is time-consuming, expensive, and repetitive. Automating the information extraction process has the potential to considerably reduce manual human labor, allowing people to focus on more creative and strategic tasks.

\begin{figure}[tbh]
    \centering
    \includegraphics[width=\textwidth]{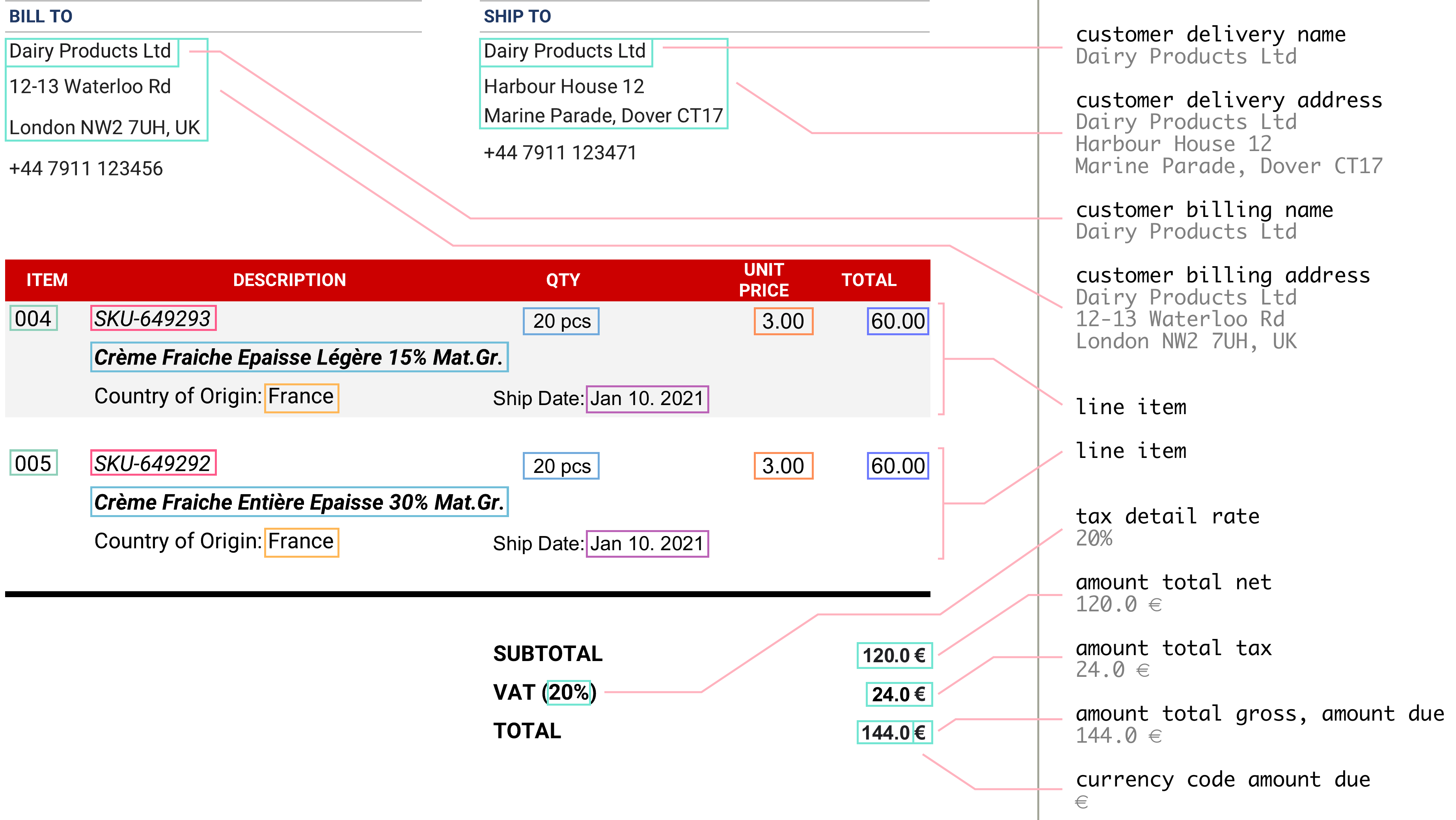}
  	\caption{Example invoice with annotations of fields and line items (LIs). Categories of fields within the LI are depicted by their color. Values of fields in the LI are not visualized in the Figure, but they are annotated in the dataset.}
  	\label{fig:invoice}
\vspace*{-2mm}
\end{figure}

Automating document information extraction is challenging because semantic and syntactic understanding is required. These documents are designed to be interpretable by humans, not machines. An example with semantic information is shown in Figure \ref{fig:invoice}. Information extraction approaches must handle varying layouts, semantic fields and multiple input modalities at the intersection of computer vision, natural language processing and information retrieval. While there has been progress on the task \cite{DBLP:conf/emnlp/KattiRGBBHF18,garncarek2021lambert,lin2021vibertgrid,denk2019bertgrid,holt2018extracting,liu2016unstructured,DBLP:journals/corr/abs-1903-12363,schuster2013intellix}, there is no publicly available large-scale benchmark to train and compare these approaches, an issue that has been noted by several authors \cite{palm2017cloudscan,sunder2019one,dhakal2019one,krieger2021information,skalicky2022business}. Existing approaches are trained on privately collected datasets, hindering their reproducibility, fair comparisons and tracking field progression~\cite{palm2019attend,palm2017cloudscan,hamdi2021information}. 

To mitigate the aforementioned issues, the DocILE lab provides a public research benchmark on a large-scale dataset. This benchmark was built by knowing the domain- and task-specific aspects of business document information localization and extraction. The DocILE benchmark will allow cross-evaluation and enable the reproducibility of experiments in business document information extraction. The dataset is the largest public source of densely annotated business documents. It consists of $8715$ annotated pages of $6680$ real business documents along with $100,000$ synthetic documents and $3.4$ million unlabeled pages of nearly a million real business documents. To mimic the real world use case, the dataset emphasizes layout diversity and contains over a thousand unique layouts. With the large amount of diverse documents and high-quality annotations, the dataset will allow researchers to investigate different aspects of document information extraction, including supervised, semi-supervised and unsupervised learning and domain adaptation.

\section{Dataset and Tasks}

\noindent
The DocILE benchmark comes with a \textit{labeled} dataset of 6680 documents from publicly available sources which were manually annotated for the tasks of \textit{Key Information Localization and Extraction} and \textit{Line Item Recognition}, described below in Sections \ref{sec:track1_kile} and \ref{sec:track2_lir} respectively. Additionally, we provide a set of 100K {\it synthetic} documents generated with the task annotations and 932K {\it unlabeled} documents, as both synthetic training data \cite{gupta2016synthetic,abs-2203-01017,dosovitskiy2015flownet} and unsupervised pre-training \cite{abs-2104-08836} have demonstrated to aid machine learning in different domains.

\subsection{Dataset Characteristics}

Table~\ref{tab:dataset-size} shows the size of the challenge dataset. All documents in the dataset were classified\footnote{Using a proprietary document type classifier from Rossum.ai.} as invoice-like documents (i.e., tax invoice, order, proforma invoice) by a model pre-trained on a private dataset. Additionally, in the labelling process, documents misclassified as invoice-like were manually removed from the dataset (e.g., budgets or financial reports, as such document types contain different information than standard invoice-like documents).

To ensure a high variance of document layouts in the dataset, {\it unlabeled} documents were clustered into layouts\footnote{We loosely define layout as the positioning of fields of each type in a document. Rather than requiring absolute positions to match perfectly, we allow transformations caused by different length of values, translations of whole sections (e.g. vertical shift caused by different lengths of tables) and translation, rotation and scaling of the whole document.}. Only a limited number of documents per layout were selected for annotation. The clustering is based on the location of field detections\footnote{The distance used for clustering relates to the difference in the relative $x$-translations between pairs of fields within a document. Vertical shifts are not penalized, since they commonly appear among documents of the same layout.} predicted by a proprietary model for KILE pre-trained on a private dataset. Furthermore, to encourage solutions that generalize well to previously unseen layouts, the train./val./test split is done such that the validation and test sets contain layouts unseen in the training set (to measure the model's generalization) as well as some seen layouts (in practice, it is common to observe known layouts and important to read them out perfectly). Meta-information describing the layouts is included in the dataset annotations. The \textit{synthetic} documents were generated using an unpublished rule-based document synthesizer based on layout annotations of 100 documents from the \textit{labeled} set.

The dataset will be shared in the form of pre-processed\footnote{Pre-processing consists of correcting page orientation, fixing or discarding broken pdfs and of de-skewing scanned documents and normalizing them to 150 DPI.} document PDFs with task annotations in JSON. As an additional resource, we will also provide predictions of text tokens (using OCR) including the location and text of the detected tokens. 

The data was sourced from two public data sources: UCSF Industry Documents Library~\cite{IndustryDocumentsLibrary} and Public Inspection Files (PIF)~\cite{PublicInspectionFiles}. The UCSF Industry Documents Library contains documents from industries that influence public health, such as tobacco companies. The majority of the documents are from the 20th century. This source was previously used to create document datasets: RVL-CDIP~\cite{harley2015icdar} (subset of IIT-CDIP~\cite{lewis2006building} and superset of FUNSD~\cite{jaume2019}) and DocVQA~\cite{MathewKJ21}. Filters in the UCSF public API \cite{IndustryDocumentsLibraryAPI} were used to retrieve only publicly available invoice-like documents with at most 3 pages, no redacted information and a threshold on document date\footnote{Old documents from this source are not included, since e.g. typewriter documents differ from today's document distribution.}. PIF contains documents (invoices, orders, "contracts") from TV and radio stations for political campaign ads. This source was previously used to create Deepform~\cite{deepform2020}.

\begin{table}[t]
\centering
\begin{tabular}{@{}lrrr@{}}
\toprule
                        & \textbf{labeled} & \textbf{synthetic}    & \textbf{unlabeled}    \\ \midrule
\textbf{documents}      & 6680          & 100 000               & 932 467               \\
\textbf{pages}          & 8715          & 100 000               & 3.4M                  \\
\textbf{layout clusters}& 1152          & 100                   & \textit{Unknown}      \\
\textbf{pages per doc.} & 1–3           & 1                     & 1-884                 \\ \bottomrule
\end{tabular}
\caption{Overview of the three parts of the challenge dataset.}
\label{tab:dataset-size}
\vspace{-6mm}
\end{table}

\subsection{Track 1: Key Information Localization and Extraction}
\label{sec:track1_kile}

The goal of the first track is to localize key information of pre-defined categories (field types) in the document. It is derived from the task of \textit{Key Information Localization and Extraction} (KILE), as defined in \cite{skalicky2022business}.

KILE extends the common definition of Key Information Extraction (KIE) by additionally requiring the location of the extracted information within the document. Such annotation is missing even in the KIE datasets~\cite{stanislawek2021kleister,borchmann2021due}. While localization is typically not needed at the end of document processing, it plays a vital role in applications that require human validation, and it is a valuable form of supervision for vision-based methods. 
Compared to \textit{Semantic Entity Recognition}, as defined by \cite{abs-2104-08836}, bounding boxes in KILE are not limited to individual words (tokens).

We focus the challenge on detecting semantically important values corresponding to tens of different field types rather than fine-tuning the underlying text recognition. Towards this focus, we provide word-level text detections for each document, we choose an evaluation metric (below) that doesn't pay attention to the text recognition part, and we simplify the task in the challenge by only requiring correct localization of the values in the documents in the primary metric. Text extractions are checked besides the locations and field types in a~separate evaluation (the leaderboard ranking does not depend on it) and any post-processing of values (deduplication, converting dates to a standardized format etc.) that is otherwise needed in practice is omitted.
With the simplifications, the main task can also be viewed as a detection problem.

\begin{figure}[tb]
    \centering
    \includegraphics[width=0.7\textwidth]{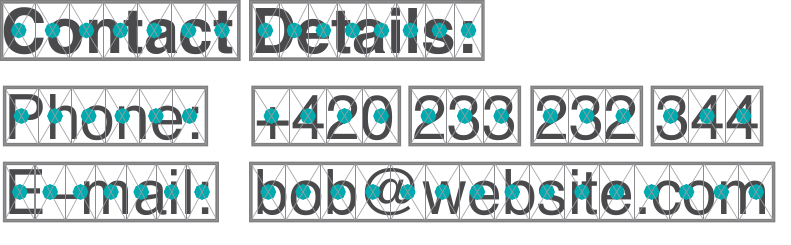}
  	\caption{Each word is split uniformly into pseudo-character boxes based on the number of characters. Pseudo-Character Centers are the centers of these boxes.}
  	\label{fig:pcc}
\end{figure}

\begin{figure}[tb]
\centering
\begin{subfigure}{.5\linewidth}
  \centering
  \includegraphics[height=50px]{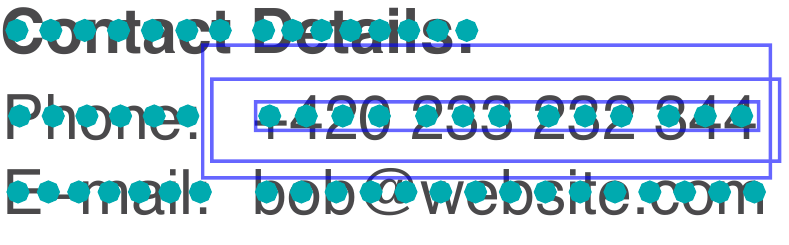}
  \caption{Correct extraction examples.}
  \label{fig:kile-metric-correct}
\end{subfigure}%
\begin{subfigure}{.5\linewidth}
  \centering
  \includegraphics[height=50px]{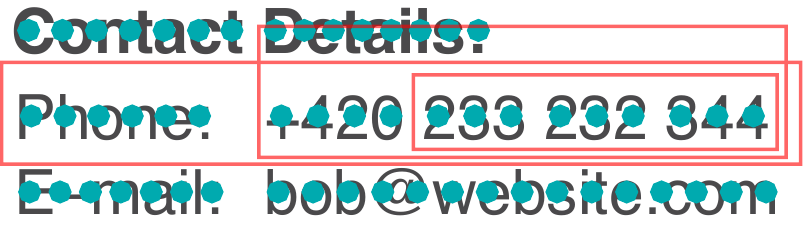}
  \caption{Incorrect extraction examples.}
  \label{fig:kile-metric-incorrect}
\end{subfigure}
\caption{Visualization of correct and incorrect bounding box predictions to capture the email address. Bounding box must include exactly the Pseudo-Character Centers that lie within the ground truth annotation. Note: In \ref{fig:kile-metric-correct}, only one of the predictions would be considered correct if all three boxes were predicted.}
\label{fig:kile-metrics-compare}
\end{figure}

\vspace{-10pt}
\subsubsection{Challenge Evaluation Metric:}
Since the task is framed as a detection problem, the standard \emph{Average Precision} metric will be used as the main evaluation metric. Unlike the common practice in object detection, where true positives are determined by thresholding the Intersection-over-Union, we use a different criterion tailored to better evaluate the usefulness of detections for text read-out. Inspired by the CLEval metric \cite{baek2020cleval} used in text detection, we measure whether the predicted area contains all related character centers (and none others). Since the character-level annotations are hard to obtain, we use CLEval's definition of Pseudo-Character Center (PCC) (see Figure~\ref{fig:pcc}). 
See Figure~\ref{fig:kile-metrics-compare} for examples of correct and incorrect detections.

Beyond the challenge leaderboard based on the metric described above, we set up a secondary benchmark for end-to-end KILE, where a correctly recognized field also needs to exactly read out the text. We invite all participants to provide the text value predictions, but it is not required for challenge submissions.

\subsection{Track 2: Line Item Recognition}
\label{sec:track2_lir}

The goal of the second track is to localize key information of pre-defined categories (field types) and group it into line items~\cite{denk2019bertgrid,holevcek2019table,palm2017cloudscan,MajumderPTWZN20,BenschPS21}. A \textit{Line Item} (LI) is a tuple of fields (i.e., \textit{description}, \textit{quantity}, and \textit{price}) describing a single object instance to be extracted, e.g., a row in a table, as visualized in Figure \ref{fig:invoice}.

This track is derived from the task of
\textit{Line Item Recognition} (LIR) \cite{skalicky2022business} and is related to \textit{Table Understanding} \cite{holevcek2019table} and \textit{Table Extraction} \cite{GobelHOO13,ZhengB0ZW21} --- problems where the tabular structure is also crucial for IE. Unlike these tasks, LIR does not explicitly rely on the structure but rather reflects the information to be extracted and stored.

\vspace{-10pt}
\subsubsection{Challenge Evaluation Metric:}
The main evaluation metric is the micro F1 score over all line item fields. A predicted line item field is correct if it fulfills the requirements from Track~1 (on field type and location) and if it is assigned to the correct line item. Since the matching of ground truth (GT) and predicted line items may not be straightforward due to errors in prediction, our evaluation metric chooses the best matching in two steps:
\begin{enumerate}[topsep=1pt]
    \item For each pair of predicted and GT line items, the predicted fields are evaluated as in Track 1.
    \item Find the maximum matching between predicted and GT line items, maximizing the overall recall.
\end{enumerate}
Similarly to the previous track, an out-of-competition end-to-end benchmark will assess the correctness of the extracted text values.

\section{Conclusions}
The first edition of the DocILE lab at CLEF 2023 and the ICDAR 2023 Competition on Document Information Localization and Extraction will present the largest benchmark for information extraction from semi-structured business documents, and will consist of two tasks: \textit{Key Information Localization and Extraction (KILE)} and \textit{Line Item Recognition (LIR)}. Participants will be given a collection of thousands of labeled documents, together with a hundred thousand of synthetic documents and nearly a million unlabeled real documents that can be used for unsupervised pre-training.

This Teaser paper summarizes the motivation and the main characteristics of the tasks. Given the input documents are practically a combination of visual- and text- inputs, we are looking forward to the contributions of several communities, including \textit{Information Retrieval}, \textit{Natural Language Processing}, and \textit{Computer Vision}.

To access the data, the repository, baseline implementations, and updates regarding the challenge, please refer to \url{https://docile.rossum.ai/}.
\newpage
\bibliographystyle{splncs04}
\bibliography{bibliography}

\end{document}